\documentclass[letterpaper, 10 pt, conference]{ieeeconf}  

\IEEEoverridecommandlockouts                              

\usepackage{times}

\usepackage{times}

\usepackage{amssymb}
\usepackage{pifont}
\usepackage{soul}
\usepackage{bbm}
\usepackage{booktabs}
\usepackage{vcell}
\usepackage{subcaption}
\usepackage{graphicx}
\usepackage{amsmath}
\usepackage{float}
\usepackage{amssymb}
\usepackage{tikz}
\usepackage{xspace}
\usepackage{bm}
\usepackage{multirow}
\usepackage{algorithm} 
\usepackage{algpseudocode} 
\usepackage{subfiles}
\usepackage{gensymb}

\newcommand{\todo}[1]{\textcolor{black}{{ }}}

\newcommand{\SE}{\mathrm{SE}}

\DeclareMathOperator{\key}{key}
\DeclareMathOperator{\query}{query}

\title{\LARGE \bf
Coarse-to-Fine 3D Keyframe Transporter
}

\author{Xupeng Zhu$^{1}$, David Klee*$^{1}$, Dian Wang*$^{1}$, Boce Hu$^{1}$, Haojie Huang$^{1}$, Arsh Tangri$^{1}$, \\ Robin Walters$^{1, 2}$, Robert Platt$^{1, 2}$ \\
\small{$^{1}$Northeastern University $^{2}$Boston Dynamics AI Institute}
\thanks{* denotes equal contribution.}
}

\begin{document}

\maketitle
\thispagestyle{empty}
\pagestyle{empty}

\begin{abstract}

Recent advances in Keyframe Imitation Learning (IL) have enabled learning-based agents to solve a diverse range of manipulation tasks.  However, most approaches ignore the rich symmetries in the problem setting and, as a consequence, are sample-inefficient. This work identifies and utilizes the bi-equivariant symmetry within Keyframe IL to design a policy that generalizes to transformations of both the workspace and the objects grasped by the gripper. We make two main contributions: First, we analyze the bi-equivariance properties of the keyframe action scheme and propose a Keyframe Transporter derived from the Transporter Networks, which evaluates actions using cross-correlation between the features of the grasped object and the features of the scene. Second, we propose a computationally efficient coarse-to-fine $\SE(3)$ action evaluation scheme for reasoning the intertwined translation and rotation action. The resulting method outperforms strong Keyframe IL baselines by an average of $>10\%$ on a wide range of simulation tasks, and by an average of $55\%$ in 4 physical experiments.

\end{abstract}

\section{Introduction}


Imitation Learning (IL) has emerged as an important approach for manipulation tasks. 
IL trains a neural network on human demonstrations to map sensory inputs into robot actions. Such demonstrations are usually limited, 
since collecting demonstrations by hand is expensive. To improve sample efficiency, Keyframe IL mimics a few keyframe gripper poses in the demonstration trajectory instead of mimicking the entire trajectory. Despite the improvement, Keyframe IL ignores the geometric structures in the manipulation policy.

Current research efforts ~\cite{james2022coarse, shridhar2023perceiver,rvt,gervet2023act3d} on Keyframe IL utilize the expressiveness of Transformer~\cite{vaswani2017attention} to infer translational actions and employed Multilayer Perceptrons (MLPs)~\cite{popescu2009multilayer} to evaluate discretized Euler angle for rotation actions. However, these design choices destroy the symmetries in the policy. Transformers are not translationally equivariant due to positional embeddings and fail to enforce locality due to the global attention mechanism. On the other hand, the Euler angle representation suffers from the discontinuity or the gimbal lock issue~\cite{5D_SO3}.

This work exploits the geometric structure of Keyframe IL to design a more sample-efficient method. We first identify a generalized form of bi-equivariant symmetry in the Keyframe IL, which extends beyond the place bi-equivariance discussed in prior works~\cite{zeng2021transporter,Huang-RSS-22,ryu2023equivariant} wherein the desired policy generalizes to independent changes in both the pick and place poses. To incorporate this property, we propose a Keyframe IL method via cross-correlation. Essentially, we derived from the place module of the Transporter Networks~\cite{zeng2021transporter, Huang-RSS-22}, where the pose actions are inferred by performing cross-correlation between a voxel representation of the scene and a dynamic kernel that represents the geometry of the grasped object. Nevertheless, extending the 2D cross-correlation in Transporter Networksto 3D suffers from curse-of-dimensionality. In 2D, the cross-correlation is performed on 3 dimensions (X, Y axes and planner rotation). In contrast, in 3D, it expands to 6 dimensions (X, Y, Z axes and roll, pitch, yaw angles), making direct computation infeasible.

To overcome the efficiency issue in 3D cross-correlation evaluation, this paper further proposes an $\SE(3)$ coarse-to-fine (C2F) calculation, extending previous translational-only C2F methods \cite{Gualtieri2020hierarchical,james2022coarse} to translation and rotation actions simultaneously. The proposed $\SE(3)$ C2F method begins by coarsely evaluating the translation and rotation actions, identifying the best $\SE(3)$ action. Then the method refines the evaluation by zooming into the best coarse $\SE(3)$ action and evaluating its neighboring translation and rotation actions. This hierarchical method drastically improved efficiency in 3D cross-correlation.

Our resulting model can learn a wide range of manipulation behaviors, including pushing, turning, using tools, etc. with a single unified architecture. This distinguishes it from prior bi-equivariant approaches~\cite{zeng2021transporter,Huang-RSS-22,ryu2023equivariant,yenchen2022mira,ryu2023diffusionedfs,huang2024fourier}, which are limited to pick-place tasks and require separated modules for picking and placing. Moreover, in contrast to existing Keyframe IL methods\cite{shridhar2023perceiver, rvt, gervet2023act3d} that are based on Transformers and Euler angles, our method evaluates action by cross-correlation on translation and rotation, effectively embeds bi-equivariance, and mitigates the gimbal lock issue. Our method outperforms Keyframe IL baselines by $>10\%$ on 18 RLBench simulation tasks with $100$ demonstrations and achieves a $55\%$ improvement on 4 real-world tasks when trained with only $10$ demonstrations.


\section{Related Work}

\textbf{Keyframe action scheme.}
ARM \cite{james2021arm} simplifies the trajectory of a closed-loop policy by using multiple keyframes. In this way, every state in the trajectory has the action that leads to the next keyframe pose and gripper aperture, and this is called demo augmentation in \cite{james2021arm}. The keyframe action scheme can be viewed as a policy between closed-loop control and open-loop control, allowing for diverse task learning while maintaining high sample efficiency in $Q$-learning \cite{james2021arm}. Following this idea, C2F-ARM\cite{james2022coarse} extends \cite{james2021arm} from 2D CNN to 3D CNN using a hierarchical evaluation style.
Later PerAct \cite{shridhar2023perceiver} adopts keyframe action in the context of imitation learning, and an extra binary collision avoidance action to let the agent seamlessly control the motion planner for complex tasks. 
More recent works ~\cite{shridhar2023perceiver, rvt, gervet2023act3d} employ Transformers to infer the translational actions and use discretized Euler angles for the gripper rotation. However standard Transformers\cite{vaswani2017attention} are not translationaly equivariant due to position embeddings assigning unique values to each position. The Euler angles representation suffers from the gimble lock issue~\cite{5D_SO3}. This work addresses these issues by using translational-rotational cross-correlation, which enforces translational equivariance and avoids the gimbal lock issue, thus gaining superior performance.

\textbf{Equivariance in robotic policy learning.}
The generalization ability of CNN is partly due to its nature of translational equivariance. \cite{zeng2018robotic,Morrison-RSS-18} showed that the translational equivariance of FCN can improve the learning efficiency of manipulation tasks.
Later, 2D rotataional equivariance are explored in \cite{wang2021equivariant,zhu2022grasp,Huang-RSS-22,zhu2023grasp,jia2023seil,wang2022so2equivariant, wang2022onrobot, zhao2022integrating,wangequivariant} and dramatically improved the sample efficiency. Several recent works~\cite{simeonov2022neural, huang2022edge,ryu2023equivariant,huang2024fourier,ryu2023diffusionedfs,huorbitgrasp,huang2024imagination} attempted encoding the 3D rotation symmetries into manipulation tasks. 
\cite{ryu2023equivariant,ryu2023diffusionedfs,huang2024fourier} achieve 3D pick-place bi-equivariance by using separate models for the pick and the place actions but are unable to perform keyframe actions. Furthermore, diffusion-based method \cite{ryu2023diffusionedfs} requires $600$ iterations for action inference, while Fourier-based method \cite{huang2024fourier} necessitates rotating a voxel grid $\sim 400$ times to perform 3D Fourier transform.
To the best of our knowledge, we are the first to recognize and leverage the Bi-equivariance in the keyframe action setting, allowing us to tackle a much broader range of manipulation problems beyond pick-place. Additionally, the proposed 3 levels of coarse-to-fine action evaluation enable one-shot action inference, making the approach computationally efficient during training and inference.

\textbf{Coarse-to-fine action evaluation.}
Evaluating all discretized $\SE(3)$ action candidates is expensive due to dimensionality. An effective evaluation is to follow a coarse-to-fine scheme~\cite{Gualtieri2020hierarchical, james2022coarse, gervet2023act3d} that gradually refines the translational action iteratively. Specifically, C2F-ARM \cite{james2022coarse} iteratively evaluates translation $q$ value in a finer voxel grid. RVT \cite{rvt} first evaluates the left view, the front view, and the top view $q$ value maps, then projects these maps to reconstruct the 3D translation $q$ value map. Act3D \cite{gervet2023act3d} iteratively evaluates sampled translation action candidates in the point cloud observation and then reduces the range of translation sampling range to refine action. Another stream of work \cite{wang2021policy, wang2021equivariant, zhu2022grasp} proposes to first evaluate the discretized translation action, then evaluate the discretized rotational action. While all of these works ignore the intertwining between translation and rotation action, we perform the coarse-to-fine action evaluation in translation while considering rotation.

\section{Background}

\textbf{Equivariance.} An equivariant function possesses the property that when the input is transformed, the output transforms accordingly. For instance, consider a planner equivariant grasping function $Q$ ~\cite{zhu2022grasp}, which takes the scene $s$ as input and outputs the gripper grasping location $a$. If the scene is transformed by a planar translation and rotation $g\in \SE(2)$, the gripper pose transforms accordingly:
\begin{align}
    Q(s) = a,\quad Q(gs) = ga.
\end{align}

\textbf{Keyframe Imitation Learning.} A keyframe action policy \cite{james2021arm, james2022coarse} specifies how the robot should move by defining a sequence of desired translations and rotations, i.e., $\SE(3)$ end-effector poses. When executing a keyframe action, the robot queries a collision-free motion planner to compute a trajectory that reaches the desired pose. The keyframe policy solves a task by executing a sequence of keyframe actions.
Keyframe Imitation Learning \cite{shridhar2023perceiver,rvt,gervet2023act3d} is a classification task that learns the action-value function $Q$ over a discretized action space, aiming to maximize the value for the discretized expert keyframe action $a$, given the observation $o$. The keyframe actions are extracted from expert demonstration trajectories by identifying moments when the gripper velocity is zero or its aperture changes \cite{james2021arm}. The observation, $o = \{s, p\}$, contains a scene representation $s$ (e.g., voxel or point cloud), and proprioceptive information, $p=\{\text{T}_{ee}, s_\text{open}, t\}$, where $\text{T}_{ee}$ is the gripper pose, $s_\text{open}$ is the gripper aperture, and $t$ is the time step. The action $a = \{a_\text{T}, a_\text{open}, a_\text{collide}\}$ specifies the desired $\SE(3)$ gripper pose $a_\text{T}$ with translation and rotation, the gripper open-close action  $a_\text{open}$, and a binary flag $a_\text{collide}$ to indicate whether to ignore collisions in the motion planning. Compared to higher frequency closed-loop control policies that output arm displacements, the keyframe framework significantly reduces the time horizon over which the policy must reason and thereby simplifies the policy learning problem.

\begin{figure}
    \centering
    \includegraphics[width=0.4\textwidth]{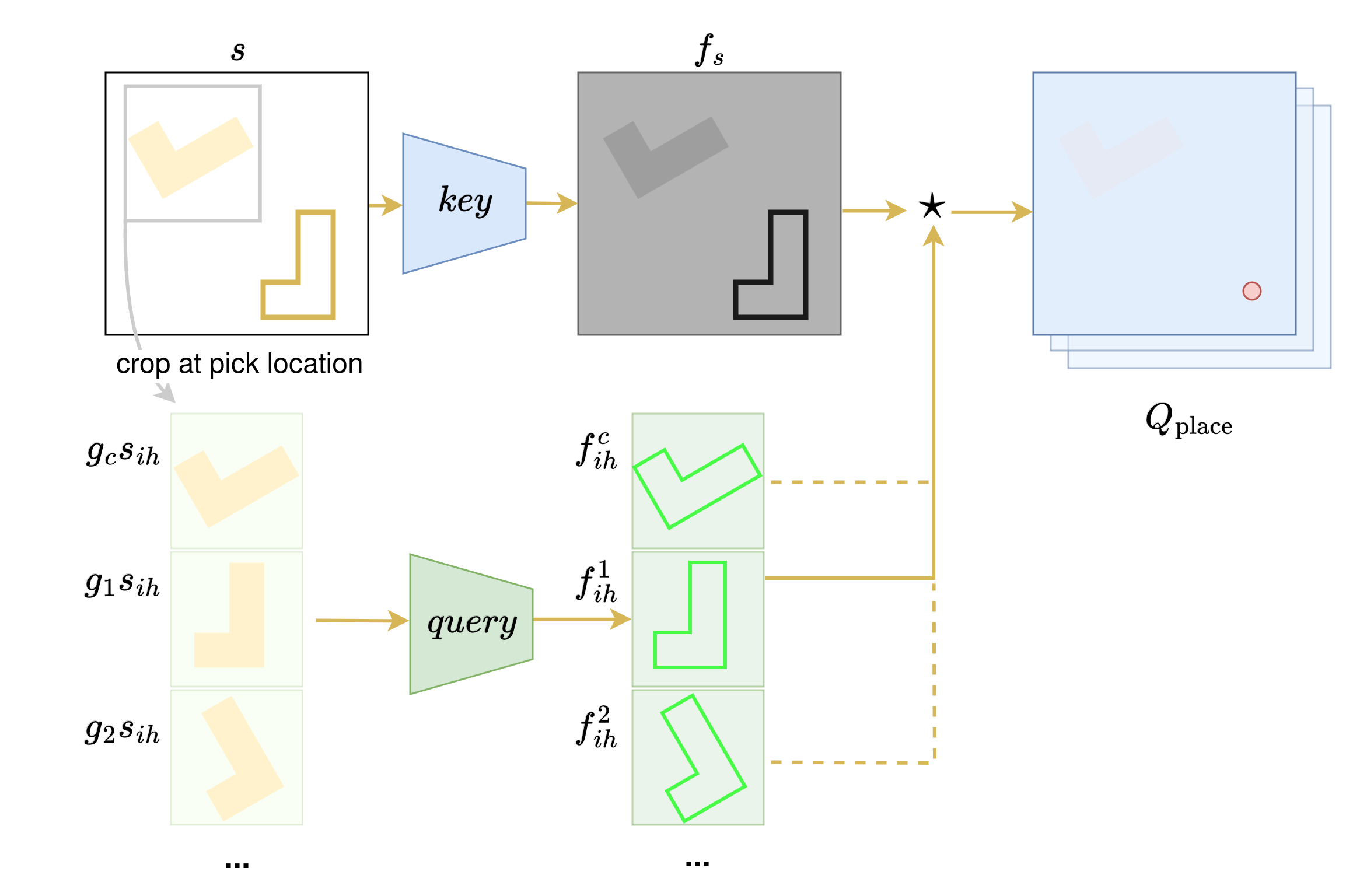}
    \caption{The place module of Transporter Networks \cite{zeng2021transporter}, along with follow-up works \cite{Huang-RSS-22, ryu2023equivariant, huang2024fourier, ryu2023equivariant} achieves bi-equivariance in place policy (e.g., picking an ``L'' shape and placing it in an ``L''-shaped receptacle) by performing cross-correlation between the scene features $f_s$ and the in-hand features $f_{ih}$. In the computed action value map $Q_\text{place}$, the height and width represent the X and Y translations of the gripper, while the channels correspond to different gripper rotations. Therefore $Q_\text{place}$ densely evaluates each trans-rotational action.}
    \label{fig:transporter}
    \vspace{-0.2cm}
\end{figure}

\textbf{The place module of Transporter Network.} The Transporter Network~\cite{zeng2021transporter, Huang-RSS-22} includes a planar pick and a planar place module that encodes rich geometric structure. The pick module is omitted for simplicity. The place module, illustrated in Figure \ref{fig:transporter}, takes the observation $s$ and the pick location $a^*_\text{pick}$ as inputs. The place module $Q_\text{place}(s, a^*_\text{pick})$ crops the observation at the pick location as the in-hand observation: $s_{ih}=crop(s, a^*_\text{pick})$. Then both the scene observation and the in-hand observation are embedded into deep latent features, maintaining the same spatial resolution, through a $\key$ and a $\query$ Unet network $f_s = \key(s), f_{ih}^i = \query(g_i s_{ih}), i \in \{1,2,...,n\}$. $f_s$ is the scene feature containing the receptacle, and $[f_{ih}^i]$ is a stack of in-hand features corresponding to each possible rotation action $g_i = \frac{2\pi i}{c}$, produced by passing rotated versions of the in-hand object observation to the $\query$ network. The place action value $Q_\text{place}$ is the result of $2$D cross-correlation in $\SE(2)$ action space between the scene features $f_s$ and each rotated in-hand feature $f_{ih}^i$, $Q_\text{place}^i = f_s \star f_{ih}^i$. The place action $a_\text{place}\in \SE(2)$ is the argmax over the place action value, $a_\text{place} = \pi(s,s_{ih}) = \arg\max Q_\text{place}(s,s_{ih})$. As \cite{Huang-RSS-22} states, if $\key$ and $\query$ networks are equivariant, then $Q_\text{place}$ is bi-equivariant due to the bi-equivariant properties of cross-correlation. However, the original Transporter Network is incompatible with keyframe action because it uses separate, specialized networks for inferring the pick action and the place action, which are coupled in a hard-coded inference sequence.

\section{Method}

\begin{figure}\centering
        \includegraphics[height=3.5cm]{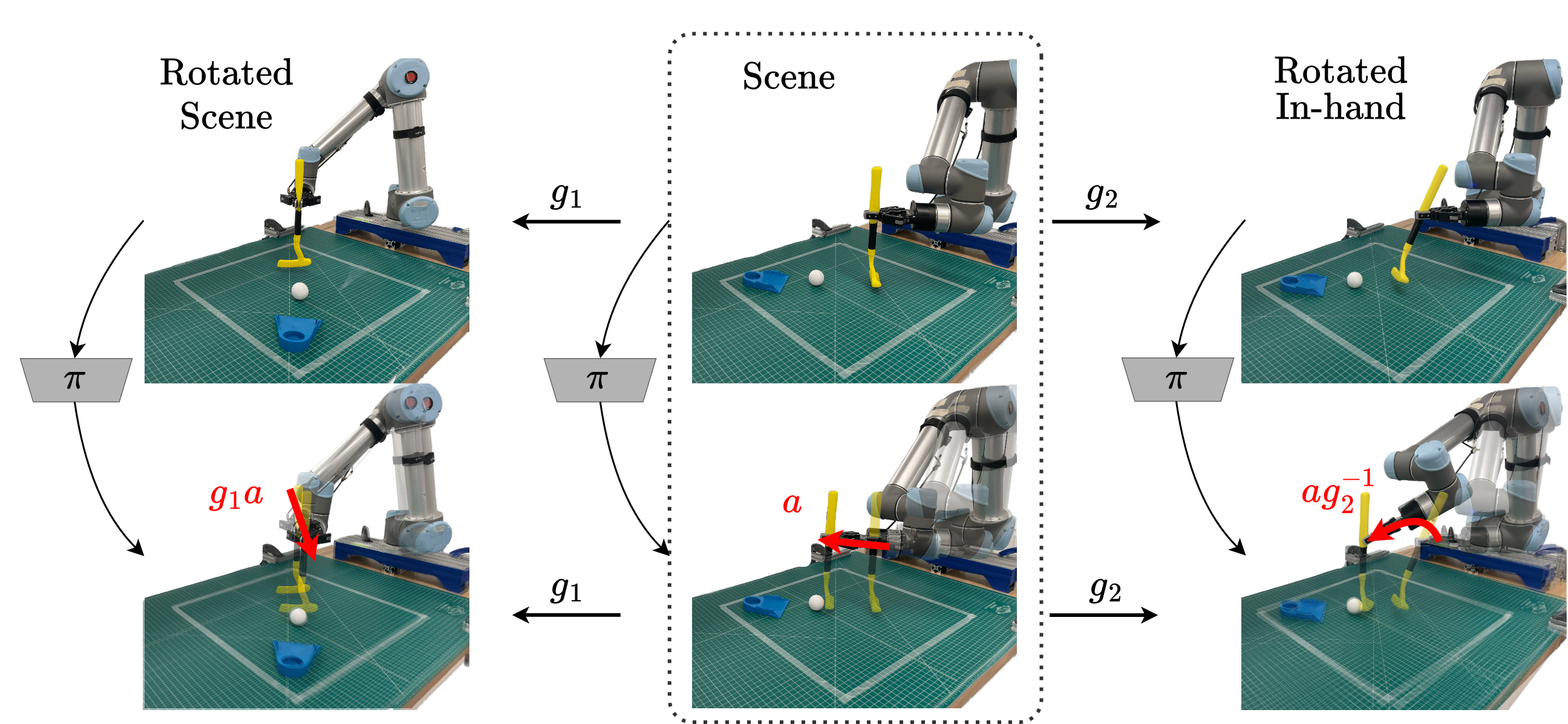}
        \caption{Bi-equivariance in keyframe policies. Second column: given a scene, the policy $\pi$ prescribes an optimal action $a$. First column: if the scene is rotated by $g_1$, the optimal action should also be rotated: $g_1a$. Third column: if the in-hand object is rotated by $g_2$, the optimal action should pre-rotate to compensate: $ag_2^{-1}$.
        }
        \vspace{-0.2cm}
\label{fig:bi_equ}
\end{figure}

\subsection{Bi-equivariance of keyframe policy.}

We find that the keyframe action policy exhibits bi-equivariance with respect to the pose action on both the scene and the in-hand objects (see Figure \ref{fig:bi_equ}). Consider the keyframe policy, denoted by the simplified notation $\pi(s,s_{ih})=a^*_\text{T}$ which takes as input the scene observation $s$ in the world frame and the in-hand observation $s_{ih}$ (the object held by the gripper) in the gripper frame, and outputs the keyframe pose action $a^*_\text{T}$. The first equivariance is when the scene is transformed by $g_1\in\SE(3)$, the pose action should be transformed by,
\begin{align}
    g_1 a^*_\text{T} = \pi( g_1 \cdot s, s_{ih})
    \label{equ:bi_equ1}
\end{align}
The second equivariance is when the grasped object is transformed by $g_2\in\SE(3)$, the pose action should compensate for this transformation by inversely transforming by $g_2^{-1}$,
\begin{align}
    a^*_\text{T}g_2^{-1} = \pi(s, g_2 \cdot s_{ih})
    \label{equ:bi_equ2}
\end{align}

Moreover, when there is grasped object(s), both Equations \ref{equ:bi_equ1} and \ref{equ:bi_equ2} are satisfied, defining what term as bi-equivariant actions, e.g., placing and using tools. When the action depends solely on the gripper, the bi-equivariance of the policy degenerates to equivariance due to the fixed gripper pose ($g_2$ becomes identity). We refer to this type of action as equivariant actions, e.g., grasping and pushing. This framework unifies the previously separate concepts of pick equivariance \cite{zhu2022grasp,zhu2023grasp,huang2022edge} and place bi-equivariance~\cite{Huang-RSS-22, huang2024fourier, ryu2023equivariant, ryu2023diffusionedfs} under a cohesive keyframe action scheme.

\begin{figure*}\centering
    \includegraphics[width=0.95\textwidth]{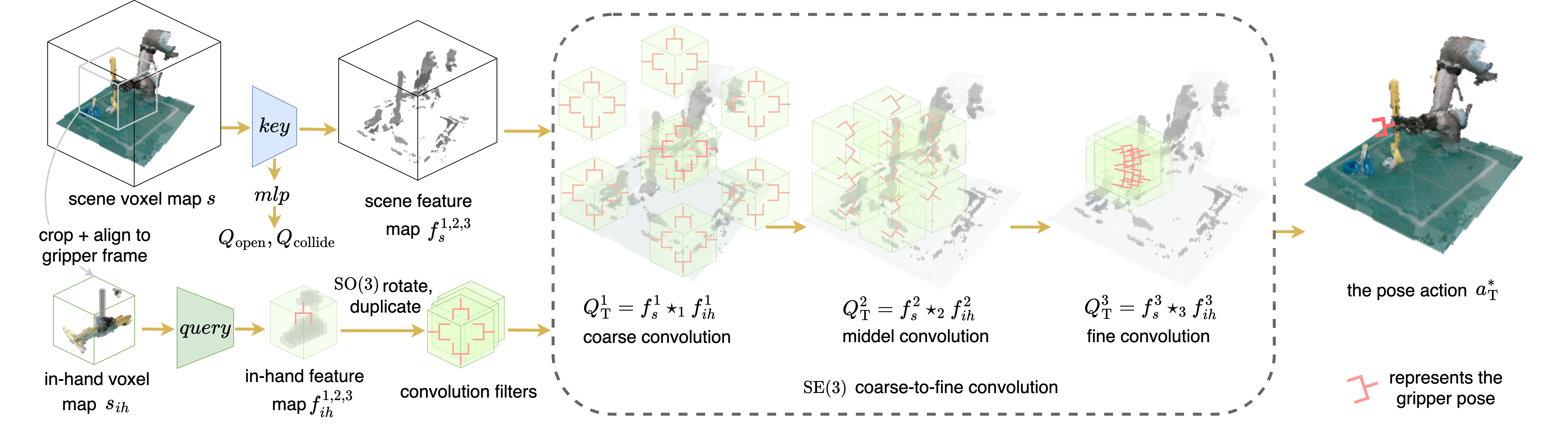}
    \caption{\textbf{Coarse-to-Fine $3$D Keyframe Transporter} inferences in two steps. Left: in step 1, the in-hand features $s_{ih}$ are obtained by cropping and transforming the scene features $s$ into the gripper frame. Then the $\key$ and $\query$ U-net networks map observations $s$ and $s_{ih}$ into pyramids of latent features $f_s^l$ and $f_{ih}^l$ respectively. Middle: in step 2, the action values $Q_\text{T}^l: \hat{G}_l \rightarrow \mathrm{R}$ are computed through a coarse-to-fine cross-correlation between the latent scene features $f_s^l$ and in-hand features $f_{ih}$. At the coarse level, the evaluated actions cover a wide translational-rotational range in a coarse grid. In the end, the fine level narrows the trans-roto range but provides fine resolution for precise action evaluation.
    Lastly, gripper open-close and planner collision actions are evaluated by MLP with the features from the $key$ U-net.}
    \label{fig:c2f_bi_equ}
    \vspace{-0.2cm}
\end{figure*}

\subsection{Keyframe IL via $3$D Cross-Correlation}
\label{sec:place_only}

We propose Keyframe IL via $3$D cross-correlation, inspired by the $2$D Transporter Net~\cite{zeng2021transporter}, to capture rich geometric structures of bi-equivariance. Several modifications are made to incorporate Transporter into the keyframe action scheme, as detailed below.

One modification is the use of the place module while discarding the pick module. Keyframe actions can be viewed as a special case of placing where anything held by the gripper is considered the in-hand object being placed, and the target pose is treated as the receptacle. This framework allows the in-hand object to be the gripper itself when no object is being grasped. Consequently, actions that rely solely on the gripper (e.g., grasping, pushing), can be interpreted as placing the gripper onto the target object. We represent the in-hand object $s_{ih}$ by canonicalizing (aligning) the scene voxel map $s$ to the gripper (end-effector) frame: $s_{ih} = crop(\text{T}_\text{ee}^{-1}\cdot s)$. 
If the in-hand observation consists only of the canonicalized gripper, as in the case of equivariant actions (e.g., picking), the proposed bi-equivariant module naturally simplifies to a single equivariance. Conversely, when the in-hand observation includes an object, the bi-equivariance property remains intact. This adaptability ensures compatibility with the keyframe bi-equivariance.

Another modification is the adaptation of the $2$D place module to a $3$D setting. To do so, the $\key$ and $\query$ Unets are replaced with $3$D Unet\cite{Unet, 3DUnet}, and the action value becomes the result of $3$D cross-correlation between scene features and in-hand features for each discretized pose action $a_\text{T}\in\SE(3)$.

However, extending the place module of the Transporter Network to $3$D poses significant computational challenges. While $2$D cross-correlation operates across 3 dimensions (X, Y axes, and planar rotation), our method performs $3$D cross-correlation across 6 dimensions (X, Y, Z axes and roll, pitch, yaw angles). This results in an exponentially increased computational cost.



\subsection{$\SE(3)$ Coarse-to-Fine Action Evaluation}
\label{sec: cross_correlation}


We present an $\SE(3)$ coarse-to-fine cross-correlation approach, extending the translational coarse-to-fine methods \cite{james2022coarse,Gualtieri2020hierarchical,gervet2023act3d} to encompass both translation and rotation. This method drastically reduces the computational complexity while maintaining high resolution in action evaluation. Additionally, we address the gimbal lock issue in \cite{shridhar2023perceiver, rvt, gervet2023act3d} by directly rotating the in-hand features multiple times to represent rotation actions. Specifically, the method first coarsely evaluates the $\SE(3)$ pose action space to identify the best coarse action. It then refines this action by zooming into the neighborhood of the coarse action and performing a finer evaluation. By repeating this process up to $l$ levels, the final level achieves high-resolution action evaluation with a significantly reduced compute.

Defining the lift cross-correlation between an input function $b$ and a dynamic filter $k$ under a group $G$ by,
\begin{align}
    (b \star k)[g] = \int_{x\in X}b(x)(g\cdot k)(x)dx, \forall g \in G,
\end{align}
where $X$ is the domain of $b$ and $k$, i.e., X, Y, and Z dimensions in our case. In practice, $X$ is represented as a voxel grid, and the group $G$ is approximated by a discrete group $\hat{G}$, which includes a translation grid along the XYZ axes times a rotation grid over the row, pitch, and yaw axes. The term $(g\cdot k)(x)$ translates and rotates the dynamic filter $k$ by $g$, for each $g$ in the grid $\hat{G}$. Notably, while the inputs reside in $X\in\mathbb{R}^3$ (a voxel grid), the output resides in $g\in\hat{G}\subset\SE(3)$ (a voxel grid times a rotation grid). Thus this cross-correlation lifts the input signal.

As shown on the left side of Figure \ref{fig:c2f_bi_equ}, we first use a $\key$ 3D Unet\cite{Unet, 3DUnet}, based on a convolutional neural network (CNN), to embed the scene observation $s$ into a pyramid of latent features $f_s^l$ at different voxel resolution levels $l$. Then a $\query$ 3D Unet embeds the in-hand observation $s_{ih}$ into features $f^{'l}_{ih}$ and predicts a mask $Q_\text{mask}$. The mask is then applied to the in-hand features to remove noise, resulting in the final masked features: $f^{l}_{ih} = Q_\text{mask} \cdot f^{'l}_{ih}$.
To infer the action $a^{l*}_\text{T}$ at level $l$, the lift cross-correlation is computed over the set of group elements $\forall g \in \hat{G}_l$,
\begin{align}
    Q^l_\text{T}[g] &= (f_s^l \star_l f_{ih}^l)[g] = \sum_{x\in X}f_s^l(x)(g\cdot f_{ih}^l)(x)
\end{align}
where $Q^l_\text{T}$ represents the pose action value at level $l$, and the action is greedily selected by $a^{l*}_\text{T} = \arg\max(Q^l_\text{T})$. At the coarsest level ($l=1$), the group $\hat{G}_1$ coarsely discretizes the action space into a low-resolution voxel grid times a low-resolution rotation grid. For finer levels ($l>1$), $\hat{G}_l$ refines the neighborhood around the optimal action from the previous level $a^{l-1*}_\text{T}$, by dividing the voxel-rotation grid into multiple finer voxel-rotation grid. This process is illustrated in the middle of Figure \ref{fig:c2f_bi_equ}.

In practice, $Q^l_\text{T}$ is a multi-channel voxel signal, where the value at each voxel corresponds to the translational action value, and each channel represents the rotational action value. To discretize the $\SE(3)$ action space into $\hat{G}_l$, we use hierarchical voxel grids for translation and Healpix grids\cite{gorski2005healpix} for rotation. The initial voxel grid has a size of $24^3$, while the rotation grid consists of $24$ discrete rotations. At each subsequent level, a voxel-rotation grid from the previous level is divided into a finer grid of size $2^3\times8$. Using a 3-level C2F process, we evaluate $\SE(3)$ action with a final resolution equivalent to a $96^3\times36864$ grid, or $1$cm in translation and $7.5^\circ$ in rotation.

\begin{figure}
    \centering
    \vspace{-0.5cm}
    
    \begin{subfigure}[b]{0.18\textwidth}\centering\includegraphics[width=\columnwidth]{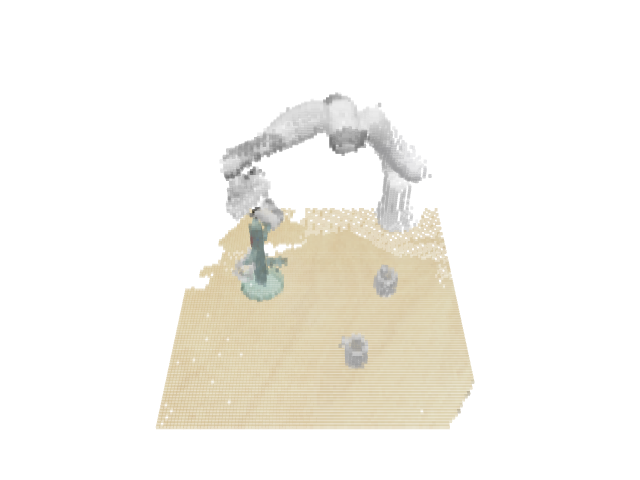}\caption{scene $s$}\end{subfigure}
    \begin{subfigure}[b]{0.14\textwidth}\centering\includegraphics[width=\columnwidth]{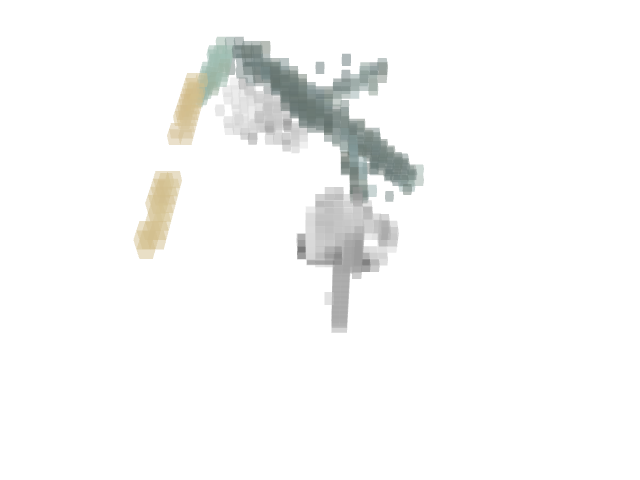}\caption{in-hand $s_{ih}$}\end{subfigure}
    \begin{subfigure}[b]{0.14\textwidth}\centering\includegraphics[width=\columnwidth]{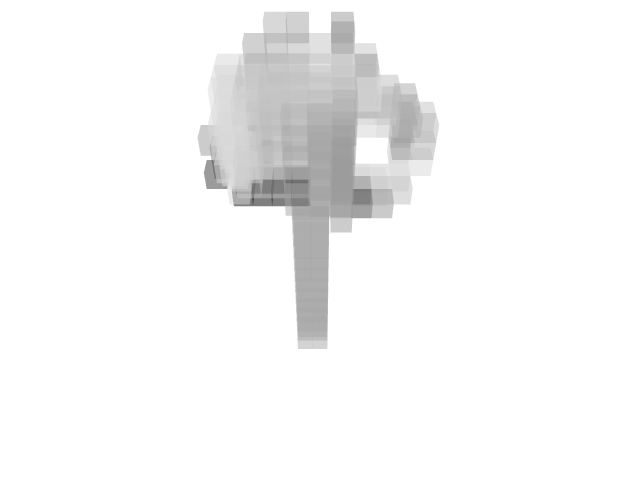}\caption{$Q_\text{mask}\cdot s_{ih}$}\end{subfigure}

    \caption{Visualization of learned in-hand segmentation.}
    
    \vspace{-0.2cm}
    \label{fig:in_hand_segmentation}
\end{figure}

\begin{table*}[!ht]
\centering
\scriptsize
\begin{tabular}{ccccccccccccccccccccc}
\toprule
       & \multicolumn{2}{c}{avg.}& \multicolumn{2}{c}{open}  & \multicolumn{2}{c}{slide}  & \multicolumn{2}{c}{sweep to} & \multicolumn{2}{c}{meat off} & \multicolumn{2}{c}{turn}  & \multicolumn{2}{c}{put in}  & \multicolumn{2}{c}{close}  & \multicolumn{2}{c}{drag}  & \multicolumn{2}{c}{stack}  \\
       & \multicolumn{2}{c}{SR$\uparrow$} & \multicolumn{2}{c}{drawer} & \multicolumn{2}{c}{block}  & \multicolumn{2}{c}{dustpan}  & \multicolumn{2}{c}{grill}    & \multicolumn{2}{c}{tap}   & \multicolumn{2}{c}{drawer}  & \multicolumn{2}{c}{jar}    & \multicolumn{2}{c}{stick} & \multicolumn{2}{c}{blocks} \\ \rule{0pt}{3ex} 
Method & 10 & 100     & 10          & 100          & 10           & 100         & 10           & 100           & 10           & 100           & 10          & 100         & 10           & 100          & 10           & 100         & 10          & 100         & 10           & 100         \\ \midrule
Ours-s  & 65 & 73 & 60        & 67           & 100          & 100         & 54           & 84            & 95           & 97            & 71          & 89          & 51           & 56           & 90           & 80          & 92          & 93          & 68           & 89            \\
RVT-s   & 62 & 61 & 98         & 94           & 89           & 93          & 60           & 40            & 77           & 96            & 89          & 96          & 37           & 42           & 81           & 84          & 99          & 95          & 25           & 31          \\
PerAct-s & 43 & 44 &89       & 93           & 100          & 100         & 1            & 0             & 98           & 98            & 83          & 77          & 19           & 28           & 56           & 73          & 21          & 30          & 85           & 59          \\
C2FARM-s & 35 & 44 &68     & 84           & 100          & 97          & 1            & 1             & 95           & 99            & 69          & 78          & 13           &  11          & 33           & 84          & 1           & 7           & 27           &  84         \\ 
\bottomrule \rule{0pt}{4ex} 
       & GPU & SGD & \multicolumn{2}{c}{screw} & \multicolumn{2}{c}{put in} & \multicolumn{2}{c}{place}    & \multicolumn{2}{c}{put in}   & \multicolumn{2}{c}{sort}  & \multicolumn{2}{c}{push}    & \multicolumn{2}{c}{insert} & \multicolumn{2}{c}{stack} & \multicolumn{2}{c}{place}  \\
       & mem $\downarrow$ & time $\downarrow$ & \multicolumn{2}{c}{bulb}   & \multicolumn{2}{c}{safe}   & \multicolumn{2}{c}{wine}     & \multicolumn{2}{c}{cupboard} & \multicolumn{2}{c}{shape} & \multicolumn{2}{c}{buttons} & \multicolumn{2}{c}{peg}    & \multicolumn{2}{c}{cups}  & \multicolumn{2}{c}{cups}   \\ \rule{0pt}{3ex} 
Method  &  &      & 10           & 100          & 10           & 100         & 10           & 100           & 10           & 100           & 10          & 100         & 10           & 100          & 10           & 100         & 10          & 100         & 10           & 100         \\ \midrule
Ours-s  & 11GB & 0.7s  & 55        & 63           & 64           & 98          & 95           & 99            & 75           & 88            & 4           & 5           & 100          & 100          & 4            & 5           & 68          & 87          & 15           & 8           \\
RVT-s  & 13GB & 0.5s & 46         & 42           & 77           & 70          & 95           & 94            & 77           & 81            & 2           & 4           & 100          & 100          & 24           & 14          & 43          & 37          & 0            & 0           \\
PerAct-s  & 13GB & 0.8s &32         & 35           & 65           & 29          & 1            & 9             & 2            & 25            & 3           & 4           & 100          & 100          & 12           & 16          & 12          & 7           & 0            & 0           \\
C2FARM-s  & 2GB & 0.1s &33       & 43           & 35           & 34          & 15           & 47            & 13           & 9             & 4           &  3          & 100          &  99          & 23           & 2           & 0           & 1           &  0           & 0           \\ 
\bottomrule
\end{tabular}
\caption{\textbf{Success rate (\%) on RLBench.} We perform a comparison between our method and various baselines on 18 RLBench tasks. The ``-s'' suffix means the agent is trained on single-task, single-variation setups. The ``-m'' suffix means the agent is trained on multi-task, multi-variation setups. Training with $10$ demos, ours achieves a similar performance to the best baseline that is trained with $\times 10$ more demos. Training with $100$ demos, ours outperforms the best baseline by $>10\%$.}
\label{table:full_rlbench18}
\vspace{-0.5cm}
\end{table*}

\subsection{In-hand segmentation}
\label{sec:in_hand_segmentation}

Bi-equivariance assumes that the in-hand object is rigidly attached to the gripper, meaning that any gripper action will transform the in-hand object identically. However, the in-hand observation could contain distracting objects that are not grasped by the gripper, which can happen, for example, when the in-hand crop size is large, and the gripper is about to grasp or release an object.

To avoid this, we propose in-hand segmentation by adding an output channel to the $\query$ network, which predicts a mask $Q_\text{mask}$ to exclude distracting objects. The method is trained in a self-supervised manner, requiring no additional labels. We compute the ground truth in-hand segmentation mask $m$ based on the observation $(s, s_{ih}, \text{T}_\text{ee})$ at time $t$ and the observation $(s', s_{ih}', \text{T}_\text{ee}')$ at time $t+1$, as well as the gripper displacement $v = \text{T}_\text{ee}^{-1}\text{T}_\text{ee}'$,
\begin{align}
    m[x] &=
    \begin{cases}
        1 & \text{if }  x \in s_{ih} > s_{ih}' + v^{-1}(s_{ih} < s_{ih}') \\
        0 & \text{if } x \in s > s' + v(s < s') \\
        -1 & \text{elsewhere}
    \end{cases}
     \nonumber 
\end{align}
where $x$ is the $XYZ-$ coordinate of the voxel grid. We use the computed segmentation mask to train the segmentation network to predict an in-hand mask. This predicted mask is then applied to filter out the features of the distracting object by performing an element-wise dot product: $f_{ih} = Q_\text{mask} f'_{ih}$, where $f'_{ih}$ represents the embedded in-hand features from the outputs of the $\query$ Unet, see Figure.\ref{fig:in_hand_segmentation} for visualizations. This approach allows us to consistently use one large in-hand crop size across all experiments.


\subsection{Bi-equivariant data augmentation}
\label{sec:data_aug}

Our method achieves discretized translational bi-equivariance through the 3D CNN backbones and approximates continuous translational and rotational bi-equivariance through bi-equivariant data augmentation. 
We augment each data point $(s, s_{ih}, a)$ in the mini-batch by applying random transformations, as described by Eqn.~\ref{equ:aug_bi_equ}, where $g_1, g_2 \in \SE(3)$ are randomly sampled,
\begin{align}
     (g_1&\cdot s, g_2\cdot s_{ih}, g_1ag_2^{-1}).
     \label{equ:aug_bi_equ}
\end{align}

\subsection{Implementation} \label{sec: implementation}
The action values $Q_\text{T}^l$ are calculated by coarse-to-fine action evaluation described in Section \ref{sec: cross_correlation}. Afterward, the agent evaluates both the gripper open action $a_\text{open}$ and the planner ignores collision action $a_\text{collide}$ using a multi-layer perceptron (MLP) based on latent features from the key network: $Q_\text{open, collide} = \text{mlp}\big(\text{maxpool}(\key(s)), \key(s)[a^*_\text{T}]\big)$, where $\text{maxpool}(\key(s))$ extracts features by maxpooling over the spatial dimension and $\key(s)[a^*_\text{T}]$ extracts features at the selected action location $a^*_\text{T}$. This MLP is similar to RVT \cite{rvt} except we do not use softmax over the feature.

The agent is trained on the expert demonstrations $\{o, a\}$ by minimizing the following loss,
\begin{align}
    \mathcal{L} = &D(Q_\text{open}, a_\text{open}) + D(Q_\text{collide}, a_\text{collide})  \nonumber\\
     & +\Sigma_{l=1}^3 D(Q_\text{T}^l, a_\text{T}^l) + \Sigma_{x}\mathbbm{1}_{m_x\ge 0}\big|\big|Q_{\text{mask}, x} - m_x\big|\big|^2_2 \nonumber
\end{align}
where $a_\text{T}^l$ is discretized expert pose action at level $l$, corresponding to the coarse-to-fine resolution $\hat{G}_l$. The indicator function $\mathbbm{1}$ equals $1$ when the mask $m_x\ge 0$, and $0$ otherwise. $Q_\text{mask}$ is the predicted in-hand segmentation mask, and $x$ refers to the position in the voxel map. We use cross-entropy loss $D$ to train action values $Q_{\{\text{T, open, collide}\}}$ and $l2$ loss $||\cdot||^2_2$ to train the in-hand segmentation mask $Q_\text{mask}$. Bi-equivariant data augmentation is applied to each sampled data point during training.


\begin{figure}\centering

    \begin{subfigure}[b]{0.48\textwidth}\centering
    \includegraphics[width=\textwidth]{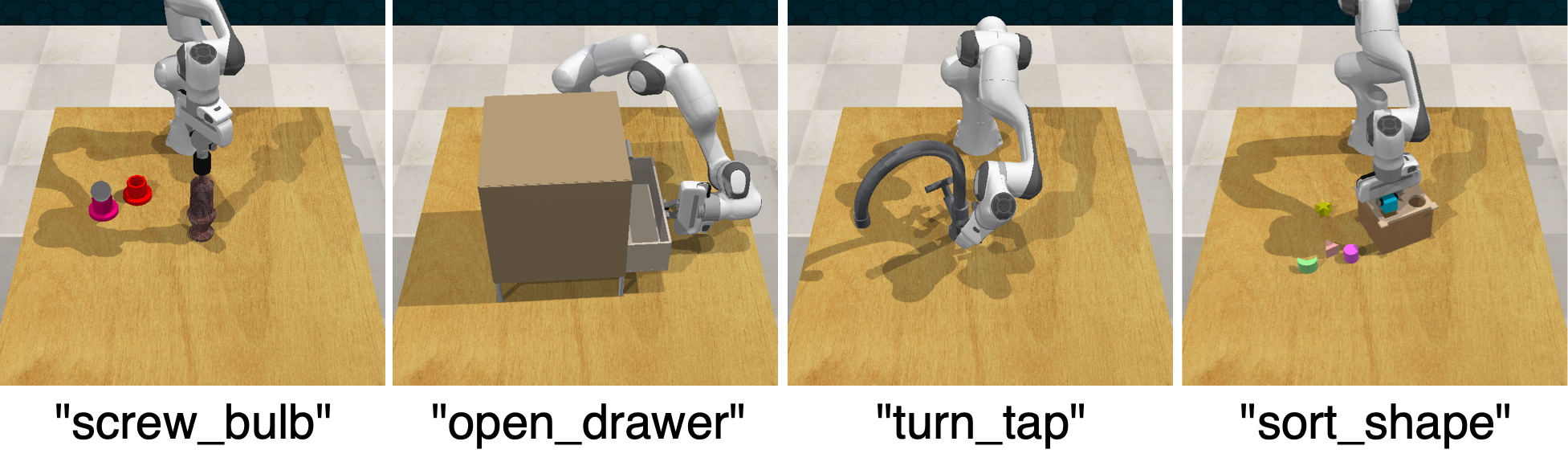}
    \caption{}
    \end{subfigure}
    \begin{subfigure}[b]{0.22\textwidth}\centering
    \includegraphics[width=\textwidth]{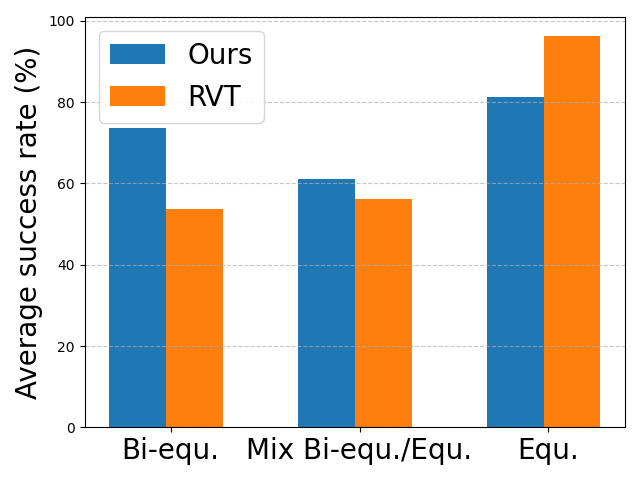}
    \caption{}
    \end{subfigure}
    \begin{subfigure}[b]{0.22\textwidth}\centering
    \includegraphics[width=\textwidth]{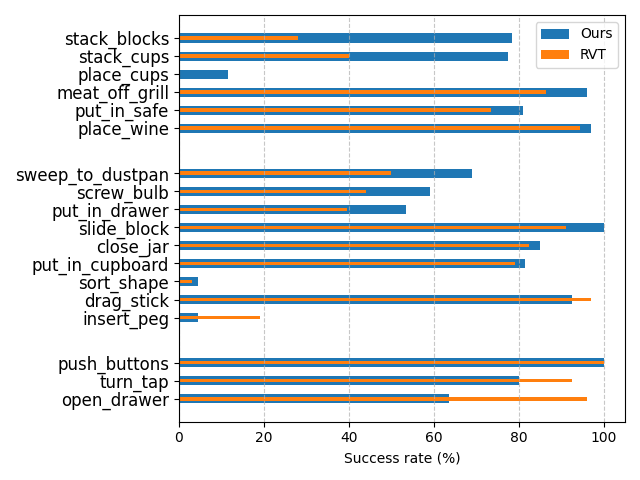}
    \caption{}
    \end{subfigure}
    \caption{(a) shows 4 out of 18 RLBench tasks~\cite{james2020rlbench}. (b) when classifying 18 tasks by the equivariance, ours has advantages on bi-equivariant and mixed equivariance tasks but underperforms RVT on equivariant tasks. (c) ``Bi-equ.'': the top 5 tasks. ``Mix Bi-equ./Equ.'': the middle 9 tasks. ``Equ.'': the button 3 tasks.
    }
    \vspace{-0.2cm}
    \label{fig:ours_vs_rvt}
\end{figure}

\section{Experiment}

\textbf{Baselines.} 
We compare our method with strong Keyframe IL baselines. Notice that we do not compare with pick-place methods\cite{zeng2021transporter, Huang-RSS-22, ryu2023equivariant, huang2024fourier} because they can not solve all 18 RLBench tasks. E.g., ``screw\_bulb" does not belong to pick-place tasks. \textbf{C2FARMBC} Coarse-to-Fine Attention-driven Robotic Manipulation Behaviour Cloning \cite{james2022coarse, shridhar2023perceiver} is an imitation learning algorithm. The method maps voxel grid input into discretized translational actions in a coarse-to-fine scheme.
\textbf{PerAct} Perceiver Actor \cite{shridhar2023perceiver} and \textbf{RVT} Robot View Transformer \cite{rvt} utilize Transformer backbones to map observation into the values of translational actions, though destroying translation equivariance. Unlike PerAct which uses expensive voxel input, RVT utilizes multi-view projected images. 
C2FARMBC employs the coarse-to-fine method but is limited to translational actions. Moreover, all of these methods represent rotation action as discretized Euler angles, which suffer from discontinuity\cite{5D_SO3}. This rotation formulation only depends on the scene observation and does not incorporate the in-hand observation.
We train all the baselines using the same parameters from the open-sourced code, except that we train on single task setup and the iteration is reduced to 15k SGD steps.

\subsection{Simulation Tasks}

\textbf{Environment:} 
We focus on the 18 tasks on the RLBench\cite{james2020rlbench}, as shown in Figure \ref{fig:ours_vs_rvt}. We generate 10 or 100 episodes of training and 100 episodes of testing demonstrations for each task. All the baselines are trained with the same training data and are tested with the same testing scenes, restored from the 100 testing demonstrations. The demonstrations include $128\times128$ RBGD camera observations from the left shoulder, right shoulder, front, and wrist camera, the proprioceptive information $p$, and the expert action $a$.

\textbf{Tasks and success metrics:} The 18 tasks are the same as \cite{shridhar2023perceiver, rvt, gervet2023act3d}, except we uses single variation. These tasks cover a wide range of manipulation policies, that includes not only pick-place (stack\_cups, stack\_blocks), but also pushing/pulling (slide\_blocks, open\_drawer), and turning (screw\_bulb, turn\_tap), etc that the pick-place methods can not solve \cite{zeng2021transporter, Huang-RSS-22, ryu2023equivariant, huang2024fourier}. The total keyframe actions in one episode for these tasks range from 2 to 14 \cite{shridhar2023perceiver}. The metric for success is binary in $\{1, 0\}$ for success or failure. The task success depends on whether the goal state is reached within 25 steps in the RLBench simulator \cite{james2020rlbench}.

\textbf{Results:} Table \ref{table:full_rlbench18} compares ours with various baselines in the 18 RLBench tasks. To make a fair comparison, we include their performance of multi-task settings as reported in the paper~\cite{shridhar2023perceiver,rvt,gervet2023act3d}. Ours outperforms all the baselines in the average success rate when trained with 10 or 100 expert demonstrations. We further analyze the performance of our method in different task groups. We first classify the 18 tasks into three categories: ``Bi-equ. tasks'': when the task mainly contains bi-equivariant actions, e.g., stack\_cups. ``Equ. tasks'': when the task only contains equivariant actions, e.g., open\_drawer. ``Mix Bi-equ./Equ. tasks'': when the task has multiple actions that contain both, e.g., put\_in\_drawer. As is shown in Figure \ref{fig:ours_vs_rvt} (b) Our method outperforms RVT in ``Bi-equ.'' and ``Mix Bi-equ./Equ.'' tasks while underperforms RVT in ``Equ.'' tasks. This indicates that our method can effectively leverage the bi-equivariant property in the task. 

\begin{figure}[!t]
    \centering
    \includegraphics[width=0.49\textwidth]{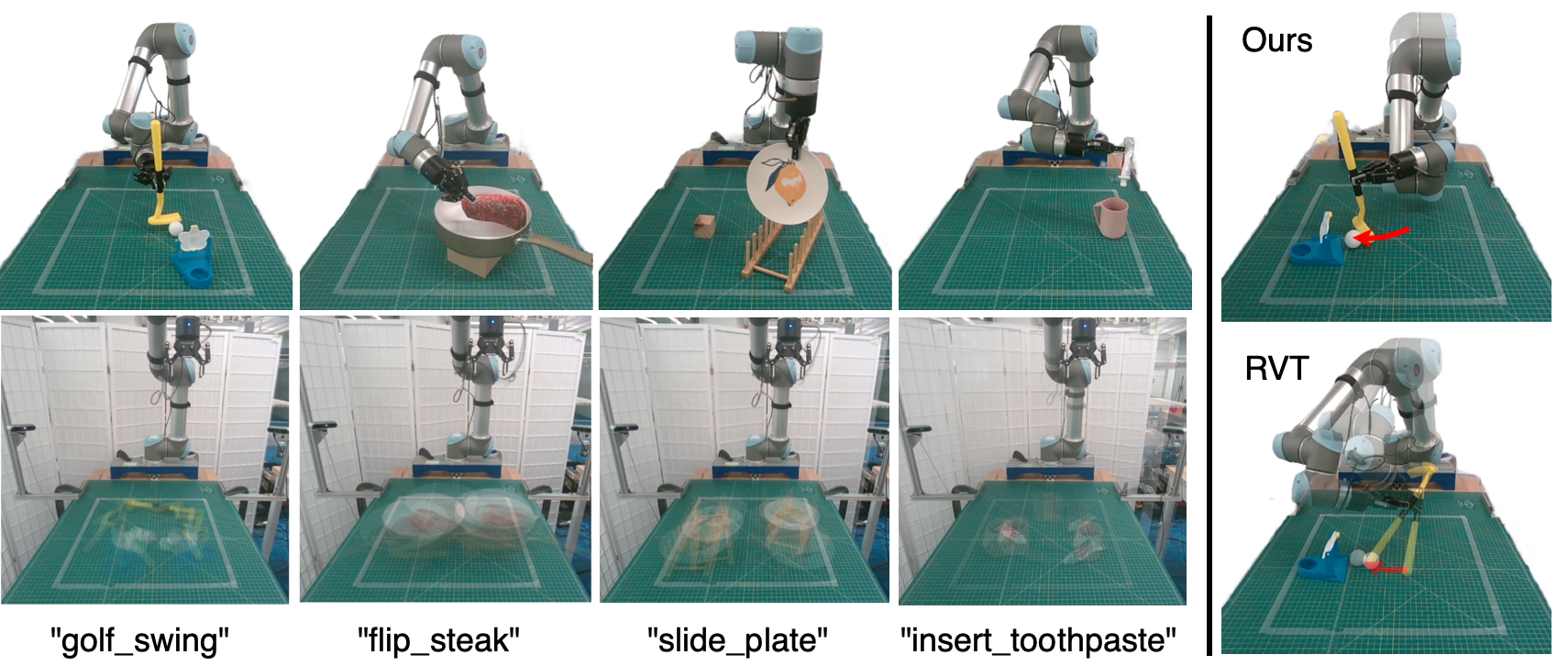}
    \caption{\textbf{Real world tasks.} \textbf{Left:} The first row shows a snapshot of 4 tasks, and the second row shows the distributions of the initial state. golf\_swing requires picking the club and aligning its head with the golf then pushing the ball to touch the goal. flip\_steak requires grasping and flipping the steak. slide\_plate requires picking the plate and then reorientating it to slide into the rack. insert\_toothpaste requires grasping the toothpaste and inserting it into the mug. \textbf{Right:} While our method is aware of the club head, RVT is not.}
    \vspace{-0.2cm}
    \label{fig:real_world_tasks}
\end{figure}

\begin{table}[!t]
\centering
\scriptsize
\begin{tabular}{ccccccccccccccccccccc}
\toprule
       & \multicolumn{2}{c}{avg.}   & \multicolumn{2}{c}{golf}   & \multicolumn{2}{c}{flip}  & \multicolumn{2}{c}{slide} & \multicolumn{2}{c}{insert}       \\
       & \multicolumn{2}{c}{SR $\uparrow$}   & \multicolumn{2}{c}{swing} & \multicolumn{2}{c}{steak}  & \multicolumn{2}{c}{plate}  & \multicolumn{2}{c}{toothpaste}  \\ \rule{0pt}{3ex} 
      & w/o        & w & w/o        & w             & w/o          & w           & w/o          & w             & w/o          & w             \\ \midrule
Ours   &77 &71 & 88         & 80            & 77           & 70          & 61           & 55            & 80           & 80            \\
RVT    &19 &16 & 14         & 10            & 44           & 40          & 17           & 15            & 0            & 0             \\ \bottomrule
\end{tabular}
\caption{\textbf{Real-world experiments with 10 training demos.} Our method outperforms RVT on all 4 tasks by an average of 55\%. ``w": the average success rate over 20 testing episodes. ``w/o": the average success rate that excludes motion planner failure.}
\label{table:real-world}
\vspace*{-0.2cm}
\end{table}

\begin{table*}[!ht]
\centering
\scriptsize
\begin{tabular}{lccccccccccccccccccccc}
\toprule
                    & avg.          & task SR      & GPU              & SGD                & stack     & sweep to  & put in    & close & drag  & screw  & put in   & place & put in      \\
Method              & SR $\uparrow$ &STD $\downarrow$& mem $\downarrow$ & time $\downarrow$  & cups      & dustpan   & drawer    & jar   & stick & bulb   & safe     & wine  & cupboard   \\ \midrule

Ours                & 83            & $\pm14$         & 11GB              & 0.7s              &  87       &  84       &  56       &  80   &  93   &  63    &  98      &  99   &  88       \\
no coarse-to-fine \footnotemark &26 & $\pm27$         & 12GB              & 2.2s              &  20       & 100       &   4       &  29   &  12   &  18    &  27      &  10   &  13       \\
no cross, C2F, seg  & 39            & $\pm27$         & 16GB              & 0.9s              &   2       & 100       &  66       &  49   &  49   &   0    &  38      &  13   &  30       \\
no segmentation     & 80            & $\pm22$         & 11GB              & 0.7s              &  68       &  95       &  27       &  90   & 100   &  65    &  94      &  94   &  89       \\
no augmentation     & 77            & $\pm13$         & 11GB              & 0.7s              &  79       &  56       &  87       &  78   &  97   &  56    &  76      &  82   &  84       \\
\bottomrule
\end{tabular}
\caption{\textbf{Ablation study.} 
\underline{avg. SR} shows the average success rate. \underline{task SR STD} shows the standard deviation of the success rate. \underline{GPU mem} and \underline{SGD time} show the GPU memory consumption and time for 1 SGD step during training.}
\vspace{-0.5cm}
\label{table:full_ablation}
\end{table*}

\subsection{Real-World Tasks}

In this section, we compare our method with RVT~\cite{rvt} in 4 complex real-world tasks (shown in Figure \ref{fig:real_world_tasks}). RVT is the best baseline in 18 RLBench tasks in Table \ref{table:full_rlbench18} when trained in the low data regimen (10 training demonstrations). The real-world tasks differ from simulation in 1) multimodal demonstrations \cite{LSTM_GMM}, 2) noisy observations \cite{zhou2024dynpoint}, and 3) limited demonstrations.

\textbf{Robot platform:} we set up the robot platform with a 6 DoFs UR5 manipulator, a Robotiq 85 gripper. The observation $\text{T}_\text{ee}, s_\text{open}$ comes from the manipulator and the gripper sensors, while the scene observation $s$ is reconstructed voxel grid from the front, the left, and the right RealSense D455 cameras. The pose action $a_\text{T}$ specifies the target pose for an off-the-shelf planner, i.e., MoveIt~\cite{moveit} motion planner with RRT-connect algorithm~\cite{rrtconnect}. We do not use the collision action $a_\text{collide}$. The Robot Operating System (ROS) is used for communication, and the workstation is equipped with a 12GB memory 2080Ti GPU. The demo is collected using an HTC VIVE controller \cite{shridhar2023perceiver}.

\textbf{Training and evaluation metrics:} For training, we first collect 10 demos for each task using the robot platform, then train our method and RVT with 15k SGD steps. The same hyper-parameters as simulations are used, except the size of the workspace is adjusted according to the robot platform. We do not cherry picking and test the last model checkpoint. We evaluate each baseline with 20 episodes. Each episode is initialized with randomized object orientation and location within the workspace, then the initialization is recorded by the cameras. We minimize the initial state between different baselines by restoring the scene to the recorded images. A task is considered a success when the success metric is achieved within 10 steps.

\textbf{Results:} Table~\ref{table:real-world} shows the evaluation results. We report two success rates, w/ means the overall success rate, and w/o means the success rate when removing the episodes that have planner failure. Our method significantly outperforms the baselines in all evaluation metrics and all 4 tasks. Training with as few as 10 demos, our method exhibits the ability to compensate for the changes of the in-hand object, e.g., correctly using the club head to hit the golf when the gripper could grasp the club with two orientations with $180^\circ$ angle (last column of Figure \ref{fig:real_world_tasks}). In contrast, RVT infers actions ignoring the in-hand state, e.g., occasionally hits the golf by the grip. We also find that the motion planner failure accounts for $10\%$ task failure. We believe this is orthogonal to our method and a better motion planner\cite{curobo_report23, zhang2019two} could effectively address this issue.

\subsection{Ablation Study}

In this section, we ablate each piece of the method to demonstrate its importance. We compare the performance on 9 RLBench tasks. All the baselines are trained with 100 demonstrations and tested with 100 episodes.

\footnotetext{The rotation grid and the in-hand size are reduced to Healpix1 and $16^3$ to match the computation overhead with ours that uses Healpix3 and $32^3$.\label{foot:changes}}

\textbf{Baselines:} \underline{no coarse-to-fine} ablates the coarse-to-fine action evaluation in Section \ref{sec: cross_correlation} by using only one level of cross-correlation instead of 3
\textsuperscript{\ref{foot:changes}}. \underline{no cross, C2F, seg} ablates the bi-equivariance of ours by removing the coarse-to-fine evaluation (Section \ref{sec: cross_correlation}), the cross-correlation (Section \ref{sec:place_only}), and the in-hand segmentation (Section \ref{sec:in_hand_segmentation}). This baseline only uses the $\key$ Unet with the same translation resolution as ours, and discretized Euler angles, which is identical to 1 level C2FARMBC. \underline{no segmentation} ablates the in-hand segmentation (Section \ref{sec:in_hand_segmentation}) by using the output of $\query$ Unet $f_{ih}'$ without the mask $Q_{mask}$. \underline{no augmentation} ablates the bi-equivariance data augmentation (Section \ref{sec:data_aug}) through training with the raw data.

\textbf{Results:} Table \ref{table:full_ablation} shows the results on the 9 tasks. When the coarse-to-fine evaluation is removed in \underline{no coarse-to-fine}, the performance is dropped by $57\%$ and the training time is tripled. This indicates that perhaps the most important piece of our method is the coarse-to-fine action inference. \underline{no cross, C2F, seg} ablates the bi-equivariance structure leading to $>40\%$ performance drop. \underline{no augmentation} shows the bi-equivariant data augmentation contributes to $6\%$ performance increment. This indicates both the bi-equivariance neural network architecture and data augmentation improve performance, while the proposed neural network architecture plays a more important role. \underline{no segmentation} shows that removing in-hand segmentation causes performance drops and a large variance, which indicates the necessity to mask out the distractors.

\section{Conclusion and Limitations}

In this paper, we propose the Coarse-to-fine 3D Keyframe Transporter that leverages the rich geometric structure in the $\SE(3)$ policy and achieves high success rates. We begin by analyzing bi-equivariance in the Keyframe IL, then introducing a $3$D cross-correlation architecture that embeds this geometric structure. Additionally, we proposed a novel coarse-to-fine evaluation to 
significantly reduce computing. Simulation experiments show that our model outperforms multiple strong baselines on 18 RLBench tasks and the physical experiments demonstrate the method can effectively learn from a few demonstrations and generalize to random initial scenes.

One limitation of our framework is the aliasing effect \cite{cesa2022a,e2cnn} of using discretized voxel features, which impacts the stability of our dynamic filter and the performance on high-precision tasks.
This issue could be mitigated by using irreducibal representations \cite{e2cnn,cesa2022a,huang2024fourier} or using point-cloud-based features \cite{qi2017pointnet++,ryu2023diffusionedfs,gervet2023act3d}. 
Another limitation is the keyframe action does not provide fine-grained control of the trajectory.
This could be addressed by using an engineered trajectory controller \cite{curobo_report23, zhang2019two}, or by combining keyframe action with closed-loop controllers \cite{xian2023chaineddiffuser, ma2024hierarchical}.







\newpage
\bibliographystyle{IEEEtran}
\bibliography{main}

\begin{thebibliography}{10}
\providecommand{\url}[1]{#1}
\csname url@samestyle\endcsname
\providecommand{\newblock}{\relax}
\providecommand{\bibinfo}[2]{#2}
\providecommand{\BIBentrySTDinterwordspacing}{\spaceskip=0pt\relax}
\providecommand{\BIBentryALTinterwordstretchfactor}{4}
\providecommand{\BIBentryALTinterwordspacing}{\spaceskip=\fontdimen2\font plus
\BIBentryALTinterwordstretchfactor\fontdimen3\font minus \fontdimen4\font\relax}
\providecommand{\BIBforeignlanguage}[2]{{%
\expandafter\ifx\csname l@#1\endcsname\relax
\typeout{** WARNING: IEEEtran.bst: No hyphenation pattern has been}%
\typeout{** loaded for the language `#1'. Using the pattern for}%
\typeout{** the default language instead.}%
\else
\language=\csname l@#1\endcsname
\fi
#2}}
\providecommand{\BIBdecl}{\relax}
\BIBdecl

\bibitem{james2022coarse}
S.~James, K.~Wada, T.~Laidlow, and A.~J. Davison, ``Coarse-to-fine q-attention: Efficient learning for visual robotic manipulation via discretisation,'' in \emph{Proceedings of the IEEE/CVF Conference on Computer Vision and Pattern Recognition}, 2022, pp. 13\,739--13\,748.

\bibitem{shridhar2023perceiver}
M.~Shridhar, L.~Manuelli, and D.~Fox, ``Perceiver-actor: A multi-task transformer for robotic manipulation,'' in \emph{Conference on Robot Learning}.\hskip 1em plus 0.5em minus 0.4em\relax PMLR, 2023, pp. 785--799.

\bibitem{rvt}
A.~Goyal, J.~Xu, Y.~Guo, V.~Blukis, Y.-W. Chao, and D.~Fox, ``Rvt: Robotic view transformer for 3d object manipulation,'' \emph{CoRL}, 2023.

\bibitem{gervet2023act3d}
T.~Gervet, Z.~Xian, N.~Gkanatsios, and K.~Fragkiadaki, ``Act3d: 3d feature field transformers for multi-task robotic manipulation,'' in \emph{Conference on Robot Learning}.\hskip 1em plus 0.5em minus 0.4em\relax PMLR, 2023, pp. 3949--3965.

\bibitem{vaswani2017attention}
A.~Vaswani, ``Attention is all you need,'' \emph{Advances in Neural Information Processing Systems}, 2017.

\bibitem{popescu2009multilayer}
M.-C. Popescu, V.~E. Balas, L.~Perescu-Popescu, and N.~Mastorakis, ``Multilayer perceptron and neural networks,'' \emph{WSEAS Transactions on Circuits and Systems}, vol.~8, no.~7, pp. 579--588, 2009.

\bibitem{5D_SO3}
Y.~Zhou, C.~Barnes, J.~Lu, J.~Yang, and H.~Li, ``On the continuity of rotation representations in neural networks,'' \emph{CoRR}, vol. abs/1812.07035, 2018.

\bibitem{zeng2021transporter}
A.~Zeng, P.~Florence, J.~Tompson, S.~Welker, J.~Chien, M.~Attarian, T.~Armstrong, I.~Krasin, D.~Duong, V.~Sindhwani \emph{et~al.}, ``Transporter networks: Rearranging the visual world for robotic manipulation,'' in \emph{Conference on Robot Learning}.\hskip 1em plus 0.5em minus 0.4em\relax PMLR, 2021, pp. 726--747.

\bibitem{Huang-RSS-22}
H.~Huang, D.~Wang, R.~Walters, and R.~Platt, ``{Equivariant Transporter Network},'' in \emph{Proceedings of Robotics: Science and Systems}, New York City, NY, USA, June 2022.

\bibitem{ryu2023equivariant}
H.~Ryu, H.~in~Lee, J.-H. Lee, and J.~Choi, ``Equivariant descriptor fields: {SE}(3)-equivariant energy-based models for end-to-end visual robotic manipulation learning,'' in \emph{The Eleventh International Conference on Learning Representations}, 2023.

\bibitem{Gualtieri2020hierarchical}
M.~Gualtieri and R.~P. Jr., ``Learning manipulation skills via hierarchical spatial attention,'' \emph{{IEEE} Trans. Robotics}, vol.~36, no.~4, pp. 1067--1078, 2020.

\bibitem{yenchen2022mira}
L.~Yen-Chen, P.~Florence, A.~Zeng, J.~T. Barron, Y.~Du, W.-C. Ma, A.~Simeonov, A.~R. Garcia, and P.~Isola, ``Mira: Mental imagery for robotic affordances,'' 2022.

\bibitem{ryu2023diffusionedfs}
H.~Ryu, J.~Kim, H.~An, J.~Chang, J.~Seo, T.~Kim, Y.~Kim, C.~Hwang, J.~Choi, and R.~Horowitz, ``Diffusion-edfs: Bi-equivariant denoising generative modeling on se(3) for visual robotic manipulation,'' 2023.

\bibitem{huang2024fourier}
H.~Huang, O.~L. Howell, D.~Wang, X.~Zhu, R.~Platt, and R.~Walters, ``Fourier transporter: Bi-equivariant robotic manipulation in 3d,'' in \emph{The Twelfth International Conference on Learning Representations}, 2024.

\bibitem{james2021arm}
S.~James and A.~J. Davison, ``Q-attention: Enabling efficient learning for vision-based robotic manipulation,'' \emph{CoRR}, 2021.

\bibitem{zeng2018robotic}
A.~Zeng, S.~Song, K.-T. Yu, E.~Donlon, F.~R. Hogan, M.~Bauza, D.~Ma, O.~Taylor, M.~Liu, E.~Romo \emph{et~al.}, ``Robotic pick-and-place of novel objects in clutter with multi-affordance grasping and cross-domain image matching,'' in \emph{2018 IEEE international conference on robotics and automation (ICRA)}.\hskip 1em plus 0.5em minus 0.4em\relax IEEE, 2018, pp. 3750--3757.

\bibitem{Morrison-RSS-18}
D.~Morrison, J.~Leitner, and P.~Corke, ``Closing the loop for robotic grasping: A real-time, generative grasp synthesis approach,'' in \emph{Proceedings of Robotics: Science and Systems}, Pittsburgh, Pennsylvania, June 2018.

\bibitem{wang2021equivariant}
D.~Wang, R.~Walters, X.~Zhu, and R.~Platt, ``Equivariant \$q\$ learning in spatial action spaces,'' in \emph{5th Annual Conference on Robot Learning}, 2021.

\bibitem{zhu2022grasp}
X.~Zhu, D.~Wang, O.~Biza, G.~Su, R.~Walters, and R.~Platt, ``Sample efficient grasp learning using equivariant models,'' \emph{Proceedings of Robotics: Science and Systems (RSS)}, 2022.

\bibitem{zhu2023grasp}
X.~Zhu, D.~Wang, G.~Su, O.~Biza, R.~Walters, and R.~Platt, ``On robot grasp learning using equivariant models,'' \emph{Autonomous Robots}, 2023.

\bibitem{jia2023seil}
M.~Jia, D.~Wang, G.~Su, D.~Klee, X.~Zhu, R.~Walters, and R.~Platt, ``Seil: Simulation-augmented equivariant imitation learning,'' in \emph{2023 IEEE International Conference on Robotics and Automation (ICRA)}.\hskip 1em plus 0.5em minus 0.4em\relax IEEE, 2023, pp. 1845--1851.

\bibitem{wang2022so2equivariant}
D.~Wang, R.~Walters, and R.~Platt, ``$\mathrm{SO}(2)$-equivariant reinforcement learning,'' in \emph{International Conference on Learning Representations}, 2022.

\bibitem{wang2022onrobot}
D.~Wang, M.~Jia, X.~Zhu, R.~Walters, and R.~Platt, ``On-robot learning with equivariant models,'' in \emph{6th Annual Conference on Robot Learning}, 2022.

\bibitem{zhao2022integrating}
L.~Zhao, X.~Zhu, L.~Kong, R.~Walters, and L.~L. Wong, ``Integrating symmetry into differentiable planning with steerable convolutions,'' \emph{arXiv preprint arXiv:2206.03674}, 2022.

\bibitem{wangequivariant}
D.~Wang, S.~Hart, D.~Surovik, T.~Kelestemur, H.~Huang, H.~Zhao, M.~Yeatman, J.~Wang, R.~Walters, and R.~Platt, ``Equivariant diffusion policy,'' in \emph{8th Annual Conference on Robot Learning}, 2024.

\bibitem{simeonov2022neural}
A.~Simeonov, Y.~Du, A.~Tagliasacchi, J.~B. Tenenbaum, A.~Rodriguez, P.~Agrawal, and V.~Sitzmann, ``Neural descriptor fields: Se (3)-equivariant object representations for manipulation,'' in \emph{2022 International Conference on Robotics and Automation (ICRA)}.\hskip 1em plus 0.5em minus 0.4em\relax IEEE, 2022, pp. 6394--6400.

\bibitem{huang2022edge}
H.~Huang, D.~Wang, X.~Zhu, R.~Walters, and R.~Platt, ``Edge grasp network: A graph-based se (3)-invariant approach to grasp detection,'' \emph{arXiv preprint arXiv:2211.00191}, 2022.

\bibitem{huorbitgrasp}
B.~Hu, X.~Zhu, D.~Wang, Z.~Dong, H.~Huang, C.~Wang, R.~Walters, and R.~Platt, ``Orbitgrasp: Se (3)-equivariant grasp learning,'' in \emph{8th Annual Conference on Robot Learning}, 2024.

\bibitem{huang2024imagination}
H.~Huang, K.~Schmeckpeper, D.~Wang, O.~Biza, Y.~Qian, H.~Liu, M.~Jia, R.~Platt, and R.~Walters, ``Imagination policy: Using generative point cloud models for learning manipulation policies,'' \emph{arXiv preprint arXiv:2406.11740}, 2024.

\bibitem{wang2021policy}
D.~Wang, C.~Kohler, and R.~Platt, ``Policy learning in se (3) action spaces,'' in \emph{Conference on Robot Learning}.\hskip 1em plus 0.5em minus 0.4em\relax PMLR, 2021, pp. 1481--1497.

\bibitem{Unet}
O.~Ronneberger, P.~Fischer, and T.~Brox, ``U-net: Convolutional networks for biomedical image segmentation,'' \emph{CoRR}, vol. abs/1505.04597, 2015.

\bibitem{3DUnet}
{\"{O}}.~{\c{C}}i{\c{c}}ek, A.~Abdulkadir, S.~S. Lienkamp, T.~Brox, and O.~Ronneberger, ``3d u-net: Learning dense volumetric segmentation from sparse annotation,'' \emph{CoRR}, vol. abs/1606.06650, 2016.

\bibitem{gorski2005healpix}
K.~M. Gorski, E.~Hivon, A.~J. Banday, B.~D. Wandelt, F.~K. Hansen, M.~Reinecke, and M.~Bartelmann, ``Healpix: A framework for high-resolution discretization and fast analysis of data distributed on the sphere,'' \emph{The Astrophysical Journal}, vol. 622, no.~2, p. 759, 2005.

\bibitem{james2020rlbench}
S.~James, Z.~Ma, D.~R. Arrojo, and A.~J. Davison, ``Rlbench: The robot learning benchmark \& learning environment,'' \emph{IEEE Robotics and Automation Letters}, vol.~5, no.~2, pp. 3019--3026, 2020.

\bibitem{LSTM_GMM}
A.~Mandlekar, D.~Xu, J.~Wong, S.~Nasiriany, C.~Wang, R.~Kulkarni, L.~Fei{-}Fei, S.~Savarese, Y.~Zhu, and R.~Mart{\'{\i}}n{-}Mart{\'{\i}}n, ``What matters in learning from offline human demonstrations for robot manipulation,'' \emph{CoRR}, vol. abs/2108.03298, 2021.

\bibitem{zhou2024dynpoint}
K.~Zhou, J.-X. Zhong, S.~Shin, K.~Lu, Y.~Yang, A.~Markham, and N.~Trigoni, ``Dynpoint: Dynamic neural point for view synthesis,'' \emph{Advances in Neural Information Processing Systems}, vol.~36, 2024.

\bibitem{moveit}
D.~Coleman, I.~A. Sucan, S.~Chitta, and N.~Correll, ``Reducing the barrier to entry of complex robotic software: a moveit! case study,'' \emph{CoRR}, vol. abs/1404.3785, 2014.

\bibitem{rrtconnect}
J.~Kuffner and S.~LaValle, ``Rrt-connect: An efficient approach to single-query path planning,'' in \emph{Proceedings 2000 ICRA. Millennium Conference. IEEE International Conference on Robotics and Automation. Symposia Proceedings (Cat. No.00CH37065)}, vol.~2, 2000, pp. 995--1001 vol.2.

\bibitem{curobo_report23}
B.~Sundaralingam, S.~K.~S. Hari, A.~Fishman, C.~Garrett, K.~V. Wyk, V.~Blukis, A.~Millane, H.~Oleynikova, A.~Handa, F.~Ramos, N.~Ratliff, and D.~Fox, ``curobo: Parallelized collision-free minimum-jerk robot motion generation,'' 2023.

\bibitem{zhang2019two}
Z.~Zhang, S.~Chen, X.~Zhu, and Z.~Yan, ``Two hybrid end-effector posture-maintaining and obstacle-limits avoidance schemes for redundant robot manipulators,'' \emph{IEEE Transactions on Industrial Informatics}, vol.~16, no.~2, pp. 754--763, 2019.

\bibitem{cesa2022a}
G.~Cesa, L.~Lang, and M.~Weiler, ``A program to build {E(N)}-equivariant steerable {CNN}s,'' in \emph{International Conference on Learning Representations}, 2022.

\bibitem{e2cnn}
M.~Weiler and G.~Cesa, ``{General E(2)-Equivariant Steerable CNNs},'' in \emph{Conference on Neural Information Processing Systems (NeurIPS)}, 2019.

\bibitem{qi2017pointnet++}
C.~R. Qi, L.~Yi, H.~Su, and L.~J. Guibas, ``Pointnet++: Deep hierarchical feature learning on point sets in a metric space,'' \emph{Advances in neural information processing systems}, vol.~30, 2017.

\bibitem{xian2023chaineddiffuser}
Z.~Xian, N.~Gkanatsios, T.~Gervet, T.-W. Ke, and K.~Fragkiadaki, ``Chaineddiffuser: Unifying trajectory diffusion and keypose prediction for robotic manipulation,'' in \emph{Conference on Robot Learning}, 2023.

\bibitem{ma2024hierarchical}
X.~Ma, S.~Patidar, I.~Haughton, and S.~James, ``Hierarchical diffusion policy for kinematics-aware multi-task robotic manipulation,'' 2024.

\end{thebibliography}

\end{document}